\documentclass[a4paper,12pt]{article}
\usepackage[utf8]{inputenc}
\usepackage[english]{babel}
\usepackage{authblk}
\usepackage{graphicx}
\usepackage[table]{xcolor} % color for text and tables
\usepackage{mathptmx}
\usepackage[singlespacing]{setspace}
\usepackage[headheight=1in,margin=1in]{geometry}
\usepackage{fancyhdr}
\usepackage{lipsum}

\definecolor{gray}{rgb}{0.9,0.9,0.9}

\makeatletter
\def\@maketitle{%
  \newpage

  \begin{center}%
  \let \footnote \thanks
    {\LARGE \@title \par}%
    \vskip 1em
    {\large \@author \par}%
    \vskip 0.5em
    {\normalsize Wellesley College, Wellesley, USA \par}%
  \end{center}%
  \par
  \vskip 0.1em}
\makeatother

\graphicspath{{images/}}

\title{Creating Targeted, Interpretable Topic Models with LLM-Generated Text Augmentation}
\author{Anna Lieb, Maneesh Arora, Eni Mustafaraj}
 
\date{}

\begin{document}

\maketitle
\thispagestyle{fancy}

\begin{center}
\newline
\end{center}

\section*{Extended Abstract}

% The abstract should outline the impact of the work, along with (if relevant) the main theoretical contribution, data and methods used, and findings. Authors are strongly encouraged to include figures and/or tables in their submission.

Making sense of natural language text at scale is a common challenge in computational social science research, especially for domains in which researchers have access to an abundance of unlabeled data but face a shortage of labeled data. Unsupervised machine learning techniques, such as topic modeling and clustering, are often used to identify latent patterns in unstructured text data in fields such as political science and sociology \cite{Nelson2020, kowsari2019, macanovic2022, grimmer2010}. These methods overcome common concerns about reproducibility and costliness involved in the labor-intensive process of human qualitative analysis. However, two major limitations of topic models are their interpretability and their practicality for answering targeted, domain-specific social science research questions. In this work, we investigate opportunities for using LLM-generated text augmentation to improve the usefulness of topic modeling output. We use a political science case study to evaluate our results in a domain-specific application. 

\textbf{Related Work.} When applying topic modeling methods to the social sciences, researchers must make subjective decisions about how to interpret an output set of keywords \cite{grimmer2013, Nelson2020}. Previous attempts to improve topic model interpretability use a variety of techniques, including incorporating document embeddings from pre-trained language models like BERT \cite{sia2020tired, grootendorst2022} and providing semi-supervised keyword seeds \cite{eshima2020keyword}. As state-of-the-art large language models (LLMs) achieve increasingly high performance on tasks that require reasoning based on natural language prompts \cite{openai2023gpt4}, social scientists are also exploring methods for LLM-powered computational content analysis, such as prompting techniques for GPT-3.5 topic generation \cite{pham2023topicgpt}. These innovations are mostly designed for and evaluated on the same general purpose pattern recognition tasks as traditional topic models, and are not suitable for exploring targeted domain-specific research questions \cite{Nelson2020, baden2022}. Their output is driven by the general goal of topic formation, and do not take into consideration underlying social science theories. Semi-supervised approaches to topic modeling, like keyword seeding, term weighting, and prior topic-word distributions \cite{eshima2020keyword, fan2019assessing, wood2017}, incorporate some domain-specific insights; however, they place a burden on researchers to derive prior knowledge representations. Additionally, researchers' expectations and biases can impact the reliability of semi-supervised results. 

\textbf{Methods.} We propose a procedure for topic modeling using LLM-generated text augmentations, which can add targeted semantic context and real-world knowledge to the latent text. This method can improve researchers' ability to answer domain-specific research questions with topic modeling. It can incorporate social science theory without topic-specific and word-specific priors, which minimizes researcher supervision and improves reliability. In this modified technique, the topic model takes LLM-generated descriptors as input text, rather than taking raw unprocessed input text. This approach is mainly applicable to topic modeling using very short text documents, such as social media posts and single-sentence text data. 

\textbf{Case study.} Our approach is motivated by applications in ongoing political science research on critical race theory (CRT) controversies in the United States. CRT controversies emerged in American ``culture war'' debates starting in September 2020, when conservative political figures criticized anti-racism curricula in K-12 schools and workplace trainings \cite{wallace-wells-2021}. To better understand the partisan debates over CRT, we conduct content analysis on a dataset of 11,704 online news headlines related to CRT collected from the GDELT (Global Data on Events, Location and Tone) Project database \cite{leetaru2013gdelt}. Specifically, our goal is to identify salient primary actors in news coverage. This can answer political science questions such as: Do news headlines frame CRT as a grassroots issue that emerged among local-level actors like students, parents, and teachers? Or does news coverage instead frame CRT as an issue driven by political elites like legislators, executives, and news pundits? Previous research in political communications suggests that these variations in news media issue framing can significantly influence public attitudes and understandings of the issue \cite{Dennis2007, iyengar1994anyone, Lecheler2015}. 

We performed topic modeling with LLM-generated text augmentations on our headlines dataset to identify categories of actors that appear in the headlines. First, we used GPT-4 to generate brief descriptions of the primary actor in each news headline.\footnote{We used the following prompt to extract actor information from each headline: \textit{What type of actor is the primary actor in this headline? Briefly describe the primary actor. If the headline doesn't reference an actor, say so. You don't need to include the headline in your response.}} This is a key step that incorporates our research goal (to identify salient primary actors) without explicitly referencing actors or keywords that we expect in the output topics. Examples of LLM-generated text augmentations are shown in Table \ref{sample-augmentation}. Based on qualitative review of a sample of actor descriptions, we found that GPT-4 generated actor descriptions were accurate, provided timely real-world information, and offered relevant details about each actor's role in the context of the headline. However, the performance of GPT-4 may vary across applications, and future work could more thoroughly evaluate the quality of text augmentations. 

Next, we used BERTopic to generate topics \cite{grootendorst2022}, using the LLM-generated actor descriptions as input documents. As a baseline for comparison, we also trained a BERTopic model using the original raw headline text as input documents. Based on our interpretation of representative keywords and documents output by BERTopic, we assigned labels to each topic in the two topic models. The augmented topic modeling output and our qualitative interpretations are displayed in Table \ref{title-description-clusters}, and the baseline results are displated in Table \ref{title-clusters}. Some similar topics are grouped together for ease of interpretation.

\textbf{Results.} We find that unsupervised topic modeling using GPT-4 blurbs (rather than unprocessed text) creates highly interpretable categories that can be used to investigate domain-specific research questions with minimal human guidance. Whereas the baseline model identifies general themes like racism, legislation, and education, the LLM-augmented model identifies specific actors like governors, legislators, teachers, and parents. Although the baseline output mentions some of these actors, they are not cleanly grouped; for example, the largest topic places school boards, a supreme court justice, parents, and teachers all into the same topic grouping (See Table \ref{title-clusters}). This type of output is less useful for targeted analysis of salient actors that appear in news coverage. Overall, this research demonstrates that LLM-generated text can augment short documents with semantic context, real-world information, and domain-specific thematic targets. Our proposed procedure for incorporating LLM-generated text augmentations in topic models can expand the utility of existing unsupervised techniques while also maintaining interpretability and reproducibility.

\bibliographystyle{plain}
\bibliography{sources.bib}

\newpage

\section*{Figures and Tables}

\begin{table}[h]
\centering
\begin{tabular}{|p{5.5cm}p{9cm}|}
    \hline
    \rowcolor{gray} \textbf{Original headline text} & \textbf{GPT-4 primary actor description} \\
    \hline
    KBJ and CRT \raggedright & The primary actor in the headline ``KBJ and CRT'' refers to Ketanji Brown Jackson (KBJ), who is a Justice on the Supreme Court of the United States. She is a legal professional and a public figure, serving in the highest court in the United States. \\
    \hline
    DeSantis Bans CRT From K-12 Classrooms Despite Absence From Curriculum \raggedright & The primary actor in the headline is DeSantis, referring to Ron DeSantis, the Governor of Florida. He is a political figure and a member of the Republican Party, known for his conservative policies and actions within the state of Florida. \\
    \hline 
    Tennessee House Passes Bill for Black History to be Taught In Fifth Through Eighth Grade Despite State Ban on Critical Race Theory Concepts \raggedright & The primary actor in this headline is the ``Tennessee House.'' The Tennessee House refers to the lower chamber of the Tennessee General Assembly, which is the state legislature of Tennessee. This body is responsible for making and passing state laws. In this context, the Tennessee House is acting in its legislative capacity to pass a bill related to educational curriculum requirements. \\
    \hline
    Critical race theory battles are waged without enough information \raggedright & The headline does not explicitly reference a specific actor. It discusses ``critical race theory battles'' in a general sense, implying that the primary actors could be individuals or groups engaged in debates or conflicts over critical race theory, but it does not specify who these actors are. \\
    \hline 
    Critical Race Theory Seeks To Tell A More Accurate Story, Supporters Say & The primary actor in the headline is ``Critical Race Theory.'' Critical Race Theory is not a person but an academic and legal framework that examines society and culture as they relate to categorizations of race, law, and power. It seeks to understand how racism is embedded within social structures and legal systems. \\
    \hline 
    
\end{tabular}
\caption{\textbf{Sample text augmentations using GPT-4 description.} GPT-4 was prompted to generate a brief description of the primary actor in each news headline. The actor descriptions provide additional real-world context and semantic information that is specifically relevant to our research questions.}
\label{sample-augmentation}
\end{table}

\begin{table}
\centering
\begin{tabular}{|p{3.2cm}p{2cm}p{9cm}|}
    \hline
    \rowcolor{gray} \textbf{Primary actor \newline interpretation} & \textbf{\# of docs \newline in topic}& \textbf{KeyBERT representation} \\
    \hline
    CRT ideology \newline (itself) & 2128 & examines society, political, activists, groups, debate, concept, opposing, topic \\
    \hline
    School \newline administration & 664 & school boards, school districts, school district, public schools, superintendent, overseeing public \\
    \hline
    Teachers & 271 & teachers, educators, employed educators, new teachers, educational roles \\
    & 167 & educators, teachers, teachers professors, educators working, responsible educating \\
    & 121 & teachers union, teachers role, federation teachers, unions organizations \\
    \hline
    Governors & 201 & governor, mississippi governor, governor political, lieutenant governor, state executive \\
    & 197 & florida governor, governor desantis, ron desantis \\
    & 144 & glenn youngkin, youngkin politician, governor glenn, virginia governor \\ 
    \hline
    Legislators & 383 & proposed legislation, legislature, legislative action, state legislature, legislators \\
     & 137 & senator, states senator, senators elected, senate senate \\
    \hline
    Parents & 538 & group parents, parents, black father, individuals children, father individual \\
    \hline
    News media & 486 & fox news, news channel, cable news, msnbc, tucker carlson, news coverage, political commentator, television host \\
    \hline
    Republicans & 442 & republican party, gop, political party, republicans \\    
    \hline
    Joe Biden & 278 & joe biden, biden administration, president joe, political figure, president \\
    \hline
    Florida & 220 & florida education, florida state, florida school \\
    \hline
    Military & 173 & military officer, military advisor, general milley, joint chiefs, chairman, secretary defense \\
    \hline
    Attorneys general & 116 & montana attorney, attorney general, attorney, missouri attorney, indiana attorney, overseeing state \\
    \hline
    Southern Baptists & 111 & baptist convention, southern baptists, baptist organization, convention sbc \\
    \hline
    No assignment \raggedright & 2795 & Outlier documents \\
     & 2132 & Rule-based exclusion from model (contains ``does not reference'' or ``does not explicitly reference'') \\
    \hline
\end{tabular}
\caption{\textbf{BERTopic results and interpretations using GPT-4 generated article descriptions.} The model considered unigrams and bigrams, with minimum topic size = 100 documents. BERTopic produced 19 topics that were interpretable and identified specific primary actors. A representative subset of the 10 KeyBERT words are shown. }
\label{title-description-clusters}
\end{table}

\begin{table}
\centering
\begin{tabular}{|p{3.2cm}p{2cm}p{9cm}|}
    \hline
    \rowcolor{gray} \textbf{Topic \newline interpretation} & \textbf{\# of docs \newline in topic}& \textbf{KeyBERT representation} \\
    \hline
     Misc. educational and political actors & 2723 & school board, school district, public schools, brown jackson, schools, parents, curriculum, teachers \\
     \hline
     Racial conflict \newline \& controversy & 1394 & racial, racism, racist, controversy, white people, debate, critics, diversity, fox news, discussion \\
     \hline
     State-specific coverage \raggedright & 221 & textbooks florida, florida classrooms, florida schools, schools desantis, florida desantis, desantis proposes \\ 
     (FL, TX, SD, VA) & 192 & texas legislature, texas schools, texas lawmakers, banning texas, approval texas, texas bill, texas public \\
     & 149 & florida bans, bans classrooms, florida education, florida news, florida classrooms, taught florida \\ 
     & 117 & dakota bans, dakota legislature, dakota noem, dakota lawmakers, dakota gov, dakota colleges, gov kristi, \\ 
     & 100 & glenn youngkin, youngkin campaign, gov youngkin, youngkin bans, youngkin virginia \\
    \hline
     Values in the classroom \raggedright & 658 & america classrooms, racism, ethnic studies, fight schools, american schools, civics education \\
     \hline 
     Passing policies & 351 & bills banning, bills ban, bill senate, mississippi senate, transparency bill, anti bill, ban teaching \\
     & 172 & bans schools, schools banning, bans teaching, black students, teachers protest \\
     \hline
     Parents & 381 & black parents, parents protest, parents fight, parents rally, parents push, parents concerned, black mother \\
    \hline
    Republicans \& Democrats \raggedright & 192 & gop reps, gop attacks, biden proposal, issue gop, gop, gop lawmakers, back biden, biden, reps push \\ 
    & 110 & biden education, schools biden, biden pushing, biden fight, biden administration, biden democrats \\
    \hline
    Teachers & 266 & pledge teach, teachers pledging, sign pledge, new teachers, teach controversial \\ 
    \hline 
    
    Military & 208 & professor defends, academy professor, defends teaching, military academies, air force, cadets, force officer \\ 
    \hline
    Protest \& elections & 134 & board races, school board, condemning racism, board candidates, meeting debate, board meeting, dismay students, questions school \\
    & 97 & letter criticism, racist, letters opposing, issue letter \\ 
    \hline
    Southern Baptists & 117 & southern baptists, baptist convention, baptist leaders \\
    \hline
    No assignment & 4122 & Outlier documents \\
    \hline
    
\end{tabular}
\caption{\textbf{BERTopic results and interpretations using raw article titles.} The model considered unigrams and bigrams, with minimum topic size = 90 documents. BERTopic produced 18 topics that were less interpretable than the GPT-4 augmented results in Table \ref{title-description-clusters}. A representative subset of the 10 KeyBERT words are shown. }
\label{title-clusters}
\end{table}

\end{document}